\documentclass[journal,twoside,web]{ieeecolor}
\pdfoutput=1
\usepackage{generic}
\usepackage{cite}
\usepackage{amsmath,amssymb,amsfonts}
\usepackage{algorithmic}
\usepackage{graphicx}
\usepackage{textcomp}
\usepackage{multirow}
\usepackage{booktabs}
\usepackage[colorlinks=true,linkcolor=black,breaklinks=true,urlcolor=blue,citecolor=black]{hyperref}
\def\BibTeX{{\rm B\kern-.05em{\sc i\kern-.025em b}\kern-.08em
    T\kern-.1667em\lower.7ex\hbox{E}\kern-.125emX}}
\begin{document}
\title{Enhancing Clinical Evidence Recommendation with Multi-Channel Heterogeneous Learning on Evidence Graphs}
\author{Maolin Luo, and Xiang Zhang
}

\maketitle

\begin{abstract}
Clinical evidence encompasses the associations and impacts between patients, interventions (such as drugs or physiotherapy), problems, and outcomes. The goal of recommending clinical evidence is to provide medical practitioners with relevant information to support their decision-making processes and to generate new evidence. Our specific task focuses on recommending evidence based on clinical problems. However, the direct connections between certain clinical problems and related evidence are often sparse, creating a challenge of link sparsity. Additionally, to recommend appropriate evidence, it is essential to jointly exploit both topological relationships among evidence and textual information describing them. To address these challenges, we define two knowledge graphs: an Evidence Co-reference Graph and an Evidence Text Graph, to represent the topological and linguistic relations among evidential elements, respectively. We also introduce a multi-channel heterogeneous learning model and a fusional attention mechanism to handle the co-reference-text heterogeneity in evidence recommendation. Our experiments demonstrate that our model outperforms state-of-the-art methods on open data.

\end{abstract}

\begin{IEEEkeywords}
Graph Attention Network, Recommendation system, Evidence-based Medicine.
\end{IEEEkeywords}

\section{Introduction}
\label{sec:introduction}
\IEEEPARstart{E}{vidence}-based medicine (EBM) is the use of current best evidence in making decisions about patient care \cite{b1}. Randomized controlled trials and meta-analyses are primary tools of EBM, providing evidence for group outcomes \cite{b2}. Incorporating external clinical evidence into clinical decision-making, referred to as EBM practice, can assist clinicians in making more informed decisions \cite{b1, b3, b5}. Despite benefits, discovering potential evidence from the vast amount available can be challenging for clinicians, and constructing clinical evidence is a time-consuming process. 

Currently, research has focused on providing assistance to clinicians in making decisions regarding drug recommendations. However, these methods only recommend drug names based on the disease, such as SafeDrug, \cite{SafeDrug} MedRec \cite{MecRec} and DRMP \cite{b8}, which may not be enough to convince clinicians and guide their decision-making process since they require more explicit clinical evidence.

\begin{figure}[!t]
\centerline{\includegraphics[width=\columnwidth]{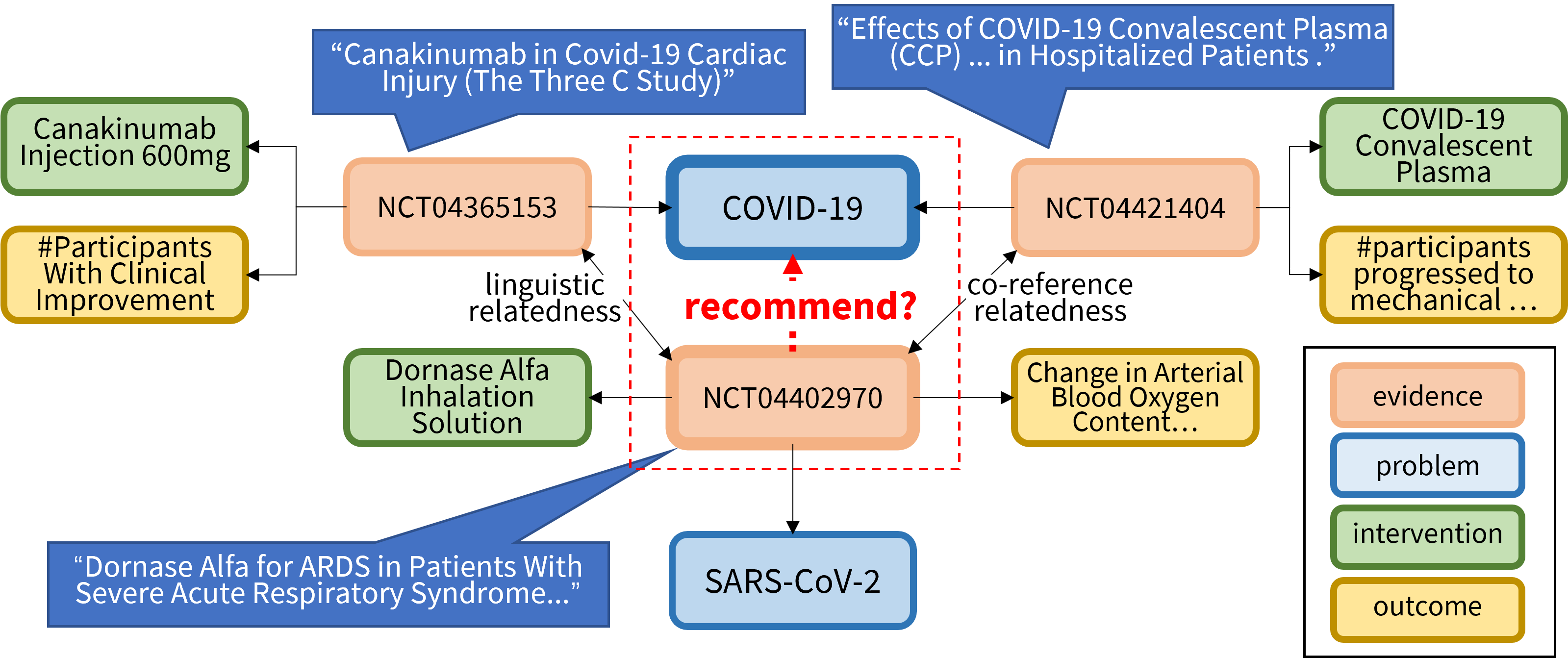}}
\caption{An example of clinical evidence recommendation. Using the available topological and textual information, we recommend clinical evidence that may be relevant to problem COVID-19 based on potential connections with previous evidence in terms of topology or textual information.}
\label{003}
\end{figure}

In this study, we utilize clinical evidence from the ClinicalTrials.gov public dataset\footnote{\href{https://clinicaltrials.gov/ct2/home}{https://clinicaltrials.gov/ct2/home}} to provide recommendations based on the potential relevance to patients' clinical problems. Each piece of evidence is described using three elements: patient/problem (P), intervention/comparison (I/C), and outcome (O), as per the PICO vocabulary\footnote{\href{https://linkeddata.cochrane.org/pico-ontology}{https://linkeddata.cochrane.org/pico-ontology}}. For instance, in Fig \ref{003}, we consider two studies (NCT04365153 and NCT04421404) focused on the problem of COVID-19, which indicate that Canakinumab Injection 600mg or therapy with COVID-19 Convalescent plasma may prevent progressive heart and respiratory failure or reduce the progression to severe hypoxemia in patients with COVID-19. There is a third study (NCT04402970) that is not designed specifically for COVID-19, but a highly-related Syndrome caused by SARS-CoV-2. It demonstrates that inhaled/nebulized dornase alfa may improve clinical outcome measures in SARS-CoV-2 related syndrome. Despite not being directly related to COVID-19, the last evidence may still be valuable for treating COVID-19 infections due to its connections with the other two studies. These connections may include linguistic similarity or co-reference to shared PICO elements, making it a good candidate for recommendation.


Our task of recommending evidence poses significant challenges. \textbf{Information sparsity} is a prevalent issue in many recommendation tasks, including those involving movies, books, music \cite{KGAT}, physical exercises \cite{b11}, and drugs \cite{SafeDrug,MecRec}. To address this challenge, we propose using the Evidence Co-reference Graph (ECG) to model the structural connections between two clinical studies that share PICO elements. However, the ECG is highly sparse, with a graph density of 0.00156. To mitigate this problem, we also introduce the Evidence Text Graph (ETG) to model the linguistic connections among studies. In addition to the challenge of information sparsity, we also need to deal with \textbf{information heterogeneity} when combining the structural and linguistic information to recommend relevant evidence for a given clinical condition. To address this challenge, we propose a Multi-Channel Heterogeneous Attention Network (MHAN) to model heterogeneous features from both evidence graphs with separate graph channels. We introduce a multi-head fusional attention in MHAN to effectively fuse the heterogeneous relevance among evidential entities. Finally, MHAN assigns a relevance score for each piece of evidence to the given problem and recommends the top-$k$ studies accordingly.




The main contributions of this paper are as follows:
\begin{itemize}
    \item We construct two clinical evidence graphs using publicly available data. The issue of graph sparsity is addressed by complementing sparse structural relevance between clinical studies with abundant linguistic relevance. 
    \item We propose a multi-head fusional attention to effectively merge heterogeneous features of evidential entities in different graphs.
    \item We propose an MHAN model to learn the relevance between evidence and problems, and we use the relevance as indicators for top-k evidence recommendation. Experiments shows that our model achieves the state-of-the-art performance comparing to other strong models. 
    
\end{itemize}

The dataset and source code of MHAN are publicly available at \href{https://github.com/wds-seu/MHAN}{https://github.com/wds-seu/MHAN}.

\section{Related Work}
\label{sec:related work}

To the best of our knowledge, no previous studies have explored clinical evidence recommendation. Nevertheless, there has been considerable research conducted on recommendation systems for movies, books, music, and drugs.

In knowledge-based recommendation, collaborative filtering is widely used in recommending tasks\cite{FM2}. LightGCN \cite{LightGCN} is proposed to aggregate high-hop neighbours. ECFKG \cite{ECFKG} applies TransE \cite{TransE} to a unified graph including users, items, entities and relations, casting the recommendation task as plausibility prediction of $\textless{}u,interact,i\textgreater{}$ triplets. KGAT \cite{KGAT} uses graph embeddings of users, items, entities and relations. Although the model considers external information and uses attention mechanism to achieve better results, it does not consider the heterogeneity of the graph. Collaborative-filtering-based recommendation models, such as SimGCL \cite{SimGCL}, aim to learn more uniform representations of user items, which can implicitly mitigate popularity bias.

Knowledge graph embedding aims to project entities and relations in the knowledge graph into a low-dimensional space. Graph Neural Networks (GNNs), such as GCN\cite{GCN} and GAT\cite{GAT}, use information aggregation from nearby nodes to learn graph representations. Wang et al. \cite{AM-GCN} proposed AM-GCN to improve the embedding capability of GCN. This is because GCN may struggle to learn complex correlation information between node features and topological structures. However, in real-world applications, knowledge graphs can be heterogeneous. RGCN\cite{RGCN} is used as the graph embedding model in MedRec\cite{MecRec} which achieve the state-of-the-art performance in drug recommendation. HGT \cite{HGT} uses node-type and edge-type dependent parameters to characterize the heterogeneous attention over each edge based on meta-relation, which achieves the state-of-the-art in the mission of knowledge embedding.

Current studies on knowledge-based recommendation suffer a lot from the problem of graph sparsity, which severely reduces the effectiveness of identifying valuable items for users. Meanwhile, heterogeneous graph neural network becomes popular for its capacity to incorporate both structured and unstructured information. But the use of heterogeneous network in recommendation systems has not been extensively investigated.

\section{Method}


\subsection{Problem Statement}

We start by formulating clinical evidence and defining the task of clinical evidence recommendation. 

\subsubsection{Clinical Evidence}
The term "Clinical Evidence" refers to a collection of summaries that present the most effective evidence-based practices in healthcare. To create this collection, we extract all relevant evidence from ClinicalTrials.gov and compile it into a set called $E$. Each piece of evidence $e \in E$ includes a textual description as well as three elements: the clinical problem being addressed ($p$), the action taken to treat or prevent the problem ($t$), and the measurement of patient outcomes ($o$) in a clinical trial. To account for comparisons made in clinical trials, which typically involve one or more interventions, we combine the "comparison" element in the PICO vocabulary with interventions. All problems, interventions, and outcomes in ClinicalTrials.gov are organized into separate sets: $P$ for problems, $T$ for interventions, and $O$ for outcomes.

\subsubsection{Clinical Evidence Recommendation}
The objective of the recommendation task is to compute a score, denoted by $\operatorname{score}(p,e)$, for each $e \in E$ with respect to a clinical problem $p$. The clinical evidence recommendation task involves identifying the top-$k$ entities, $(e_1,e_2,...,e_k)$, with the highest scores $\operatorname{score}(p,e_i)$ for a given clinical problem $p$. These top entities are regarded as highly relevant to current studies on $p$, and therefore, are useful for medical practitioners who are interested in $p$.

\subsection{Construction of Evidence Graphs}

To encompass both the structural and linguistic associations in clinical evidence, we construct two graphs of evidence: the Evidence Co-reference Graph (ECG), which centers on the shared PICO elements among clinical studies, and the Evidence Text Graph (ETG), which centers on the textual similarities among them.

\subsubsection{Evidence Co-reference Graph}

The Evidence Co-reference Graph $\mathcal{G_C} = (\mathcal{U_C},\mathcal{R_C},\gamma,\phi)$ is a heterogeneous graph constructed straightforwardly from the evidence set. $\mathcal{U_C}$ is the node set, comprising evidence set $E$ and the referred sets of $P$, $T$ and $O$. $\mathcal{R_C}$ is the edge set, in which each edge $ r\in\mathcal{R_C}=(u_i,u_j)$  connects an $u_i\in E$ to an evidence element $u_j \in P \cup T \cup O$. $\gamma$ and $\phi$ are two mapping functions, in which $\gamma(u)$ maps a node $u$ to its specific type of $E$, $P$, $T$ and $O$, and $\phi(u_i,u_j)$ maps an edge to its possible type of ``$E$-$P$", ``$E$-$T$" or ``$E$-$O$".



\subsubsection{Evidence Text Graph}

The evidence text graph $\mathcal{G_T} = (\mathcal{U_T},\mathcal{R_T})$ is a homogeneous graph. The node set $\mathcal{U_T}$ comprises all the evidence in $E$. Given an $ r\in\mathcal{R_T}=(u_i,u_j)$, it indicates that the evidence $u_i$ and $u_j$ has a high text similarity between their descriptions. We calculate the similarity of two evidential nodes based on the Euclidean distance of their BERT \cite{BERT} embeddings to reflect their latent semantic relevance.  

For each $u \in \mathcal{U_T}$, its textual description is encoded as a 768-dimensional vector $H[u]$ using BERT \cite{BERT}. Given $u_i, u_j$ $ \in \mathcal{U_T}$, the Euclidean distance between their embeddings $H[u_i]$ and $H[u_j]$ is denoted as $D(H[u_i],H[u_j])$, and their text similarity $\eta_{ij}$ is calculated as Eq.1. We set a similarity threshold $\eta_{thr}$ to filter out trivial linguistic connections: $ (u_i,u_j)\in\mathcal{R_T} \rightarrow \eta_{ij}\geq {}\eta_{thr}$

\begin{equation}
\eta_{ij} = 1-\frac{D(H[u_i],H[u_j])}{\operatorname{max}(D(H[u]))-\operatorname{min}(D(H[u]))}
\end{equation}

\subsection{Embedding of Evidence Graphs}
\begin{figure*}[!t]
\centerline{\includegraphics[width=2\columnwidth]{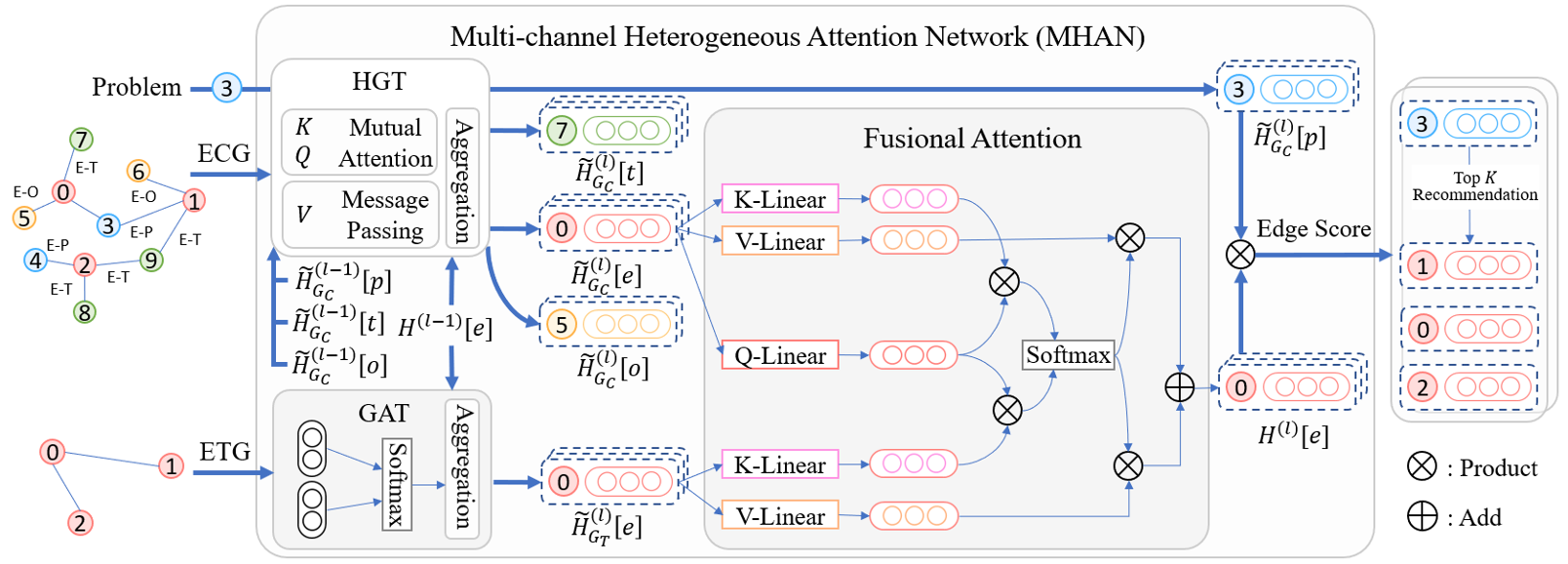}}
\caption{The architecture of Multi-Channel Heterogeneous Attention Network (MHAN). Given a problem in interest $p$, the ECG $\mathcal{G_C} = (\mathcal{U_C},\mathcal{R_C},\gamma,\phi)$, ETG $\mathcal{G}_T = (\mathcal{U_T},\mathcal{R_T})$ and the initial embedding $H^{(0)}$ of each entities in both graphs, the MHAN model takes them as input to learn a fusional topological and linguistic representation $H^{(l)}$ for each entities in ECG and ETG. MHAN includes three components: (1) ECG embedding component HGT, (2) ETG embedding component GAT, (3) a fusional attention. A relevance ranking score is computed for each problem/evidence pair and it is applied to the downstream recommendation task. In this example, the evidence 0 and 1 are recommended because they are explicitly related to the problem 3 as stated in ECG, while evidence 2 is recommended because it shares an intervention with evidence 1 and is therefore implicitly related to the problem in interest.}
\label{000}
\end{figure*}

\subsubsection{Evidence Co-reference Graph Embedding}
We use the BERT pre-training model to produce a vector $H^{(0)}[u]$ representing the textual description for each node $u \in \mathcal{U_C}$. This vector serves as the initial embedding of $u$ . To maintain dedicated representations for different types of nodes and edges, we incorporate the Heterogeneous Graph Transformer (HGT) \cite{HGT} into our model. We apply an iterative propagation method to update the node representation as follows:

\begin{equation}
   A_C = \underset{\forall u_i \in \mathcal{N}(u_j)}{\operatorname{Softmax}} \left( \left(K(u_i,\gamma(u_i)) W_{\phi(e)}^{A T T} Q(u_j, \gamma(u_j) \right) \cdot \frac{\mu}{\sqrt{d}} \right)
\end{equation}

\begin{equation}
    \widehat{H}^{(l)}_{\mathcal{G_C} }[u_j] = \underset{\forall u_i \in \mathcal{N}(u_j)}{\oplus}\left( A_C \cdot  V \left(u_i,\gamma(u_i) \right) W_{\phi(e)}^{M S G} \right)
\end{equation}

\begin{equation}
    \widetilde{H}^{(l)}_{\mathcal{G_C} }[u_j] = \sigma \left( \operatorname{Linear}_{\gamma(u_j) }\widehat{H}^{(l)}_{\mathcal{G_C} }[u_j] \right) + H^{(l-1)}[u_j]
\end{equation}

The linear projections $K(\cdot),Q(\cdot),V(\cdot)$ are derived from node embedding $H^{(l-1)}$, indexed by the node type $\gamma(u)$. For each edge type $\phi(r)$, the matrices $W_{\phi(r)}^{A T T}$ and $W_{\phi(r)}^{M S G}$ are unique edge-based matrices. The tensor $\mu$ captures the relational triplet $<\gamma (u_i),\phi (r), \gamma (u_j)>$, and $d$ denotes the dimension of the vectors $H[u]$. By applying softmax, the node attention $A_C$ of $\mathcal{G_C}$ is computed, followed by message passing that updates the embedding of the target node from ${H}^{(l-1)}_{\mathcal{G_C}}[u_j]$ to $\widetilde{H}^{(l)}_{\mathcal{G_C}}[u_j]$. $\mathcal{N}(u_j)$ represents all the neighboring nodes that point to the target node $u_j$.


\subsubsection{Evidence Text Graph Embedding}
The Evidence Text Graph $\mathcal{G_T}$ is represented as a homogeneous graph, where only evidence nodes $u \in E$ are included. The presence of an edge between nodes $u_i$ and $u_j$ is determined by their similarity $\eta_{ij}$ being greater than the similarity threshold $\eta_{thr}$. We initialize each node's vector with the BERT-encoded vector $H^{(0)}[u]$ that corresponds to its textual description. To learn the representation of each evidence based on its textual description and linguistic relationships, we employ the GAT\cite{GAT} as a part of our embedding layer in an iterative manner:


\begin{equation}
    A_T = \underset{\forall e_i \in N(e_j)}{\operatorname{Softmax}}\left(\vec{a}\left(W H^{(l-1)}[e_j]~ \| ~ W H^{(l-1)}[e_i]\right)\right)
\end{equation}
    
\begin{equation}
    \widetilde{H}^{(l)}_{\mathcal{G_T} }[e_j] = \sigma \left(\operatorname{Mean}\left( A_T \cdot W H^{(l-1)}[e_i] \right)\right)
\end{equation}

In order to obtain sufficient textual expressive power, we projection $H^{(l-1)}[E]$ to suitable-level features. The source node feature and the target node feature are then concatenated. $A_T$ is calculated as the node attention by softmax. We pass message from source node to target node to update the embedding of it, where $\sigma(x)=\frac{1}{\exp (-x)+1}$ is the sigmoid function.


\subsubsection{ Attention Mechanism}

A node u Although we have obtained both ECG and ETG embeddings of the Evidence graphs, nodes $E$ have different embeddings $\widetilde{H}^{(l)}_{\mathcal{G_C}}[E]$ and $\widetilde{H}^{(l)}_{\mathcal{G_T}}[E]$ in $\mathcal{G_C}$ and $\mathcal{G_T}$, respectively. We need to fuse the information of nodes $E$ in both knowledge graphs to form a unified entity representation ${H}^{(l)}[E]$. 

To merge the different embeddings of the same entity on multiple graphs, AM-GCN\cite{AM-GCN} introduced an adaptive attention mechanism, which has shown promising results in semi-supervised classification tasks. The adaptive attention learns the importance weights of embeddings in an adaptive manner. The formula for fusing the information of node $e_i$ is expressed as follows:

\begin{equation}
    A_{ad}=\operatorname{Softmax} \left( \operatorname{MLP}\left( \begin{bmatrix} \widetilde{H}^{(l)}_{\mathcal{G_C}}[e_i]~,~\widetilde{H}^{(l)}_{\mathcal{G_T}}[e_i] \end{bmatrix} \right) \right)
\end{equation}

\begin{equation}
    H^{(l)}[e_i]=A_{ad} \begin{bmatrix} \widetilde{H}^{(l)}_{\mathcal{G_C}}[e_i]~,~\widetilde{H}^{(l)}_{\mathcal{G_T}}[e_i] \end{bmatrix}
\end{equation}

The embeddings of each node $e_i$ in $\mathcal{G_C}$ and $\mathcal{G_T}$ are $\widetilde{H}^{(l)}_{\mathcal{G_C}}[e_i] \in \mathbb{R}^{d}$ and $\widetilde{H}^{(l)}_{\mathcal{G_T}}[e_i] \in \mathbb{R}^{d}$, respectively, and they have the same dimension. We stack them together to form $\begin{bmatrix} \widetilde{H}^{(l)}_{\mathcal{G_C}}[e_i]~,~\widetilde{H}^{(l)}_{\mathcal{G_T}}[e_i] \end{bmatrix} \in \mathbb{R}^{2 \times d}$. We project the stacked matrix to a matrix $\in \mathbb{R}^{2 \times 1}$ by a 2-layer multilayer perceptron whose activation function is $tanh()$. We then obtain a set of attention coefficients $A_{ad}$ for $e_i$ on $\mathcal{G_C}$ and $\mathcal{G_T}$, by calculating the softmax function. After applying the attention coefficients to $\widetilde{H}^{(l)}_{\mathcal{G_C}}[e_i] \in \mathbb{R}^{d}$ and $\widetilde{H}^{(l)}_{\mathcal{G_T}}[e_i] \in \mathbb{R}^{d}$ respectively, the summation yields a uniform vector $H^{(l)}[e_i] \in \mathbb{R}^{d}$ of $e_i$ representation.

In addition to the node fusion methods proposed in previous works, it is important to evaluate their effectiveness in downstream tasks of recommender systems.  Various fusion methods such as concatenation, summation, and averaging have been employed to combine the information from two nodes. To preserve the semantic information of the nodes, we adopt an attention-based approach where we sum the embeddings $\widetilde{H}^{(l)}_{\mathcal{G_C}}[e_i]$ and $\widetilde{H}^{(l)}_{\mathcal{G_T}}[e_i]$ with weights. To ensure the importance of each graph in the fusion process, we propose a shared matrix $W$ which is constrained between 0 and 1. For each node $e_i$, the weights of its embeddings in the two graphs are denoted by $W_i$ and $1-W_i$. The fused representation of node $e_i$ is then obtained by taking the Hadamard product of the shared matrix $W$ and the concatenated embeddings.
\begin{equation}
    H^{(l)}[e_i]= W[i] 
\odot \widetilde{H}^{(l)}_{\mathcal{G_C}}[e_i]+(1-W[i]) \odot \widetilde{H}^{(l)}_{\mathcal{G_T}}[e_i]
\end{equation}

The weight coefficient matrix $W$ provides greater flexibility to the model in learning the importance of each node embedding compared to the attention coefficients $a_{ad}$, which are dependent on the node embeddings. Our experiments demonstrate that the method achieves better results than adaptive attention on recommender system tasks.

However, calculating attention coefficients by discarding node information may lead to overfitting issues. Therefore, we still prefer to compute the attention coefficients based on the node embeddings.

In the ECG, nodes $e$ gather information from their neighboring nodes $p$, while in the ETG, nodes $e$ only gather information from their neighboring nodes $e$. For the recommender system task of recommending a set of $e$ nodes based on a $p$ node, we propose a fusional attention method. We use $p$ as a query to obtain the attention coefficients. The multi-head fusional attention mechanism is formulated as follows:

\begin{equation}
    H^{(l)}[e_i]= \underset{n=1}{\overset{N}{\big| \big|}}  ~{\operatorname{Softmax}}\left( \frac{Q^{n}(e_i)K^{n}(e_i)^T}{\sqrt{ d/N} }\right) V^{n}(e_i)
\end{equation}
\begin{gather*}
   Q^{n}(e_i)=\operatorname{Q-Linear}^{n}\left(\widetilde{H}^{(l)}_{\mathcal{G_C} }[e_i]\right) \\
K^{n}(e_i)=\operatorname{K-Linear}^{n}\left( \begin{bmatrix} \widetilde{H}^{(l)}_{\mathcal{G_C}}[e_i]~,~\widetilde{H}^{(l)}_{\mathcal{G_T}}[e_i] \end{bmatrix} \right)\\
V^{n}(e_i)=\operatorname{V-Linear}^{n}\left(\begin{bmatrix} \widetilde{H}^{(l)}_{\mathcal{G_C}}[e_i]~,~\widetilde{H}^{(l)}_{\mathcal{G_T}}[e_i] \end{bmatrix}\right)
\end{gather*}

We project the embedding $\widetilde{H}^{(l)}_{\mathcal{G_C} }[e_i]$ of node $e_i$ into the N-th $Query$ vector $Q^{n}(e_i)$ with a linear projection $\operatorname{Q-Linear}^{n}$: $\mathbb{R}^{d} \rightarrow \mathbb{R}^{\frac{d}{N}}$, where $N$ is the number of attention heads and $\frac{d}{N}$ is the vector dimension per head. Similarity, we also project the stacked matrix $\begin{bmatrix} \widetilde{H}^{(l)}_{\mathcal{G_C}}[e_i]~,~\widetilde{H}^{(l)}_{\mathcal{G_T}}[e_i] \end{bmatrix} \in \mathbb{R}^{2 \times d}$ into the N-th $Key$ matrix and $Value$ matrix with linear projection: $\mathbb{R}^{2 \times d} \rightarrow \mathbb{R}^{2 \times \frac{d}{N}}$, respectively. Next, we calculate the similarity between $Q^{n}(e_i)$ and each of the $K^{n}(e_i)$. We multiply the attention coefficient, which is calculated by the softmax function and $V^{n}(e_i)$ to get the embedding per head. Finally, we concatenate $N$ attention heads together to get the embedding of $H^{(l)}[e_i]$.

The node embedding $H^{(l)}[E]$ of all nodes $E$ and $P,T,O$ embedding in ECG contribute to the representation of all nodes in $l$-th layer together.

\begin{equation}
    H^{(l)} = \begin{Bmatrix} {H}^{(l)}[E], ~\widetilde{H}^{(l)}_{\mathcal{G_C}}[P] ,~\widetilde{H}^{(l)}_{\mathcal{G_C}}[T] ,~\widetilde{H}^{(l)}_{\mathcal{G_C}}[O]  \end{Bmatrix}
\end{equation}

\subsection{Evidence Recommendation}
We define $\operatorname{score}(p,e)$ to characterize the relevance ranking score between each $p$ and $e$. $\sigma$ is the sigmoid function. As shown in Eq. 12, given an input clinical problem $p_i$, we recommend the top $k$ entities $(e_1,e_2,...,e_k)$ with the highest relevance scores.

\begin{equation}
    score(p_i,e_i) = \sigma \left( H[p_i] \cdot H[e_i]^T \right)
\end{equation}

MHAN is trained in an end-to-end manner. Given a clinical problem $p$ that is already referred in a piece of evidence $e$, we assume that $e$ should have a high relevance score to $p$, since there is an explicit relation between them in the dataset. And the score $\operatorname{score}(p,e)$ should be higher than any score between $p$ and a set of $k$ randomly sampled evidence $\left\{e^{\prime}\right\} \in E$ that have no explicit relations to $p$. We define the margin loss as follows:

$\left\{e^{\prime}\right\}$ from an arbitrary noise distribution $e^{\prime} \sim {P_n}(e)$ that . We negatively sample $k$ $e^{\prime}$ nodes for each set of edges connecting $p$ and $e$. We define the margin loss as follows:
\begin{equation}
    \mathcal{L}=\sum_{e_{i} \sim P_{n}(e), i=1, \ldots, k} \max \left(0, 1-y_{p, e}+y_{p, e_i}\right)
    \label{eq13}
\end{equation}

\section{Experiments}

In this section, we evaluate MHAN to answer following research questions:
\begin{itemize}
    \item \textbf{RQ1: }How does MHAN perform comparing to various state-of-the-art recommendation models?(in section B)
    \item \textbf{RQ2: }How do different modules contribute to the overall model performance? (in section C)
    \item \textbf{RQ3: }Could we simply combine ECG and ETG in a single channel? (in section C)
    \item \textbf{RQ4: }How does fusional attention perform comparing to the state-of-the-art adaptive attention and other non-attention mechanisms? (in section D)
    \item \textbf{RQ5: }How does the similarity threshold of ETG affect the performance? (in section E)
    \item \textbf{RQ6: }How does MHAN perform with different fusional attention heads hyper-parameter settings? (in section G)
    \item \textbf{RQ7: }How to explain the recommendation results? (in section F)
\end{itemize}

\subsection{Experimental Setup}

\subsubsection{Datasets}

The performance of the proposed MHAN is evaluated on the clinical evidence dataset from ClinicalTrials.gov. The statistics of the dataset is summarized in Table \ref{table1}. Our dataset contains  247 latest evidence studying on 180 problem, with 361 interventions and 629 outcomes. The statistics on connections in ECG and ETG derived from this dataset is summarized in Table \ref{table2}.

\begin{table}
\caption{Statistics of our experiment dataset}
\centering
\begin{tabular}{c|ccccc}
\hline
Data              & \#p & \#i & \#o & \#e & \#edge  \\
\hline
ClinicalTrials.gov & 180 & 361 & 629 & 247 & 1563    \\
\hline
\end{tabular}
\label{table1}
\end{table}

\begin{table}
\caption{Statistics on the number of edges of each evidence graph}
\centering
\begin{tabular}{ccc|c}
\hline
\multicolumn{3}{c|}{ECG}                                                           & ETG     \\ \hline
\multicolumn{1}{c|}{hasProblem} & \multicolumn{1}{c|}{hasIntevention} & hasOutcome & related \\ \hline
\multicolumn{1}{c|}{449}        & \multicolumn{1}{c|}{485}            & 629        & 522     \\ \hline
\end{tabular}
\label{table2}
\end{table}

\begin{table*}
\centering
\caption{Experimental results of different methods on clinical evidence}
\begin{tabular}{l|cccc|cccc} 
\toprule
\multicolumn{1}{c|}{\multirow{2}{*}{Model}} & \multicolumn{8}{c}{Metrics}                                                                                                            \\
\multicolumn{1}{c|}{}                       & NDCG@3         & NDCG@5         & NDCG@7         & NDCG@9         & HR@3           & HR@5           & HR@7           & HR@9            \\ 
\hline
NFM\quad(2017)                                         & 0.102          & 0.105          & 0.135          & 0.144          & 0.111          & 0.133          & 0.200          & 0.222           \\
ECFKG\quad(2018)                                       & 0.067          & 0.075          & 0.090          & 0.103          & 0.067          & 0.089          & 0.122          & 0.156           \\
KGAT\quad(2019)                                        & 0.189          & 0.197          & 0.205          & 0.222          & 0.200          & 0.222          & 0.267          & 0.300           \\
LightGCN\quad(2020)                                    & 0.150          & 0.151          & 0.168          & 0.185          & 0.189          & 0.244          & 0.256          & 0.267           \\
SimGCL\quad(2022)                                      & 0.159          & 0.157          & 0.185          & 0.199          & 0.189          & 0.233          & 0.244          & 0.300           \\
RGCN\quad(2018)                                        & 0.247          & 0.254          & 0.260          & 0.279          & 0.244          & 0.289          & 0.322          & 0.344           \\
HGT\quad(2020)                                         & 0.277          & 0.274          & 0.280          & 0.289          & 0.233          & 0.300          & 0.333          & 0.356           \\
MHAN                                        & \textbf{0.284} & \textbf{0.319} & \textbf{0.326} & \textbf{0.325} & \textbf{0.256} & \textbf{0.344} & \textbf{0.378} & \textbf{0.400}  \\
\bottomrule
\end{tabular}
\label{table3}
\end{table*}

\subsubsection{Evaluation Metrics}
We use the general metrics for recommendation task including $HR@K$ and $nDCG@K$ ($K=3,5,7,9$).

\begin{equation}
    H R @ K=\frac{NumberofHits@K }{|G T|}
\end{equation}

The numerator of the equation represents the sum of the number of correctly recommended samples among the top $k$ recommendations while the denominator $|GT|$ represents the total number of samples in the test set.
\begin{equation}
    n D C G @ K=\frac{D C G @ K}{I D C G @ K}
\end{equation}

$nDCG@K$ aims to give higher scores to items that are ranked higher on the list while penalizing items that are ranked lower. The score for each item is calculated as its Discounted Cumulative Gain ($DCG$) up to that point, divided by the Ideal DCG ($IDCG$), which is the DCG of a perfect ranking. In other words, if the relevant evidence entities are ranked higher in the recommendation list, they receive higher scores.

\subsubsection{Training Details} 
To ensure reproducibility of our work, we set a fixed random seed of 2022. The training and test sets are split randomly with a ratio of 8:2. The MHAN model with 2 layers, a 256 hidden dimension, threshold of 0.8, and learning rate of 0.001 is trained. We use 4 heads for the HGT and GAT embedding component and 16 heads for the fusional attention. The model is implemented using python==3.9.13, pytorch-cu116 and dgl-cu116.

All initial vector of baselines are initialized with the BERT model the same as MHAN. Their initial parameters are set according to the suggested by their papers and we also further carefully turn parameters to get optimal performance.

\subsection{Comparison with Baselines}
To evaluate the performance of our method MHAN, we compare MHAN with the recent competitive state-of-the-art baseline methods including both in recommending systems and knowledge-based embedding.

\begin{itemize}
    \item \textbf{NFM}\cite{NFM} is a neural network-based recommendation model that combines the strengths of both Factorization Machines (FM) and deep learning models.
    \item \textbf{ECFKG}\cite{ECFKG} is a collaborative filtering method that uses knowledge graph embeddings to enhance the recommendation performance.
    \item \textbf{KGAT}\cite{KGAT} is a recommendation model that utilizes graph embeddings of users, items, entities and relations to capture the collaborative and content information of items by attention mechanism.
    \item \textbf{LightGCN}\cite{LightGCN} is an efficient recommendation method that can handle sparsity in the data, and incorporate the graph structure of the data to improve the quality of the recommendations.
    \item \textbf{SimGCL}\cite{SimGCL} is a comparative learning recommendation method that  learns more uniform user-item representations that can implicitly mitigate the popularity bias.
    \item \textbf{RGCN}\cite{RGCN} is a model that uses graph convolutional networks (GCN) to learn representations of entities in a graph that can handle multiple relations between entities.
    \item \textbf{HGT}\cite{HGT} is a graph embedding model that uses transformer-based architecture and multi-head attention mechanism to model the interactions between nodes in a heterogeneous graph.
\end{itemize}

The results of evidence recommendation are presented in Table \ref{table3}. In general, the collaborative filtering-based method NFM performs better on dense networks but is less effective in sparse clinical evidence recommendations. Two recommendation system methods, KGAT and ECFKG, consider external node information such as intervention entities and outcome entities. Among them, KGAT achieves better results than ECFKG by using an attention mechanism. LightGCN is designed to handle sparse graphs, while SimGCL uses a comparative learning approach. These methods outperform traditional recommender system approaches but still do not surpass KGAT as they do not incorporate external information. RGCN and HGT learn the heterogeneous graph ECT as an embedding representation and use negative sampling based on Equation \ref{eq13}. They utilize the heterogeneity of the graph and the information of all entities, with HGT achieving better results by using a transformer-based architecture and multi-head attention.

Our proposed MHAN model outperforms all the baseline models, which can be attributed to the inclusion of the ETG in our modeling approach. The linguistic links in the ETG reveal the interrelationships among evidence entities beyond mere co-reference. We use two separate channels for representation learning of the heterogeneous ECG and the homogeneous ETG. The fusion of evidence entities by multi-head fusional attention takes full account of heterogeneous information and textual information, which mitigates graph sparsity and solves co-reference-text heterogenetity.


\begin{table}
\centering
\caption{The results of variants}
\begin{tabular}{l|cccc} 
\toprule
       & REeb  & CRec & URec  & MHAN            \\ 
\hline
NDCG@3 & 0.157 & 0.277 & 0.275 & \textbf{0.284}  \\
NDCG@5 & 0.153 & 0.274 & 0.279 & \textbf{0.319}  \\
NDCG@7 & 0.153 & 0.280 & 0.283 & \textbf{0.326}  \\
NDCG@9 & 0.157 & 0.289 & 0.287 & \textbf{0.325}  \\
HR@3   & 0.144 & 0.233 & 0.233 & \textbf{0.256}  \\
HR@5   & 0.189 & 0.300 & 0.289 & \textbf{0.344}  \\
HR@7   & 0.189 & 0.333 & 0.333 & \textbf{0.378}  \\
HR@9   & 0.211 & 0.356 & 0.356 & \textbf{0.400}  \\
\bottomrule
\end{tabular}
\label{table4}
\end{table}

\subsection{Ablation Study}
The contribution of the individual components of MHAN to its functionality is discussed in this subsection. 
\subsubsection{Initial Embedding}
We construct the ECG and ETG using linguistic links but replace the BERT-based initial embedding with a random initial embedding to demonstrate the impact of heterogeneous textual information as an initial embedding for clinical evidence recommendation. As shown in Table \ref{table4}, we observe that the BERT-based embedding performs better than the random embedding (REeb), indicating that the embedding of text-based information plays a significant role in learning a superior representation of the two graphs.

\subsubsection{Multi-channel}
We define the recommendation only by ECG using only channel as CRec. We add the triples ("evidence", "similar", "evidence") from ETG to the ECG, constructing the unified evidence graph (UEG). We define the recommendation by UEG as URec.

According to Table \ref{table4}, we can observe that the results of CRec and URec are comparable, but both are lower than the results obtained through two-channel representation learning using MHAN for ECG and ETG. This shows that linguistic links are not the same as traditional entity links, as they introduce interference information while uncovering potential linguistic links and mitigating graph sparsity. MHAN, on the other hand, performs embedding through two channels, one for type heterogeneity of nodes and edges and another for text heterogeneity of linguistic links, which leads to better results.


\subsection{Study on Node Fusion }
\begin{table}
\centering
\caption{Performance of different node fusion methods}
\begin{tabular}{l|ccc} 
\toprule
       & Adaptive Attention & Shared Matrix & Fusional Attention  \\ 
\hline
NDCG@3 & 0.144              & 0.243    & \textbf{0.284}  \\
NDCG@5 & 0.155              & 0.250    & \textbf{0.319}  \\
NDCG@7 & 0.174              & 0.255    & \textbf{0.326}  \\
NDCG@9 & 0.192              & 0.263    & \textbf{0.325}  \\
HR@3   & 0.122              & 0.233    & \textbf{0.256}  \\
HR@5   & 0.144              & 0.278    & \textbf{0.344}  \\
HR@7   & 0.178              & 0.322    & \textbf{0.378}  \\
HR@9   & 0.211              & 0.333    & \textbf{0.400}  \\
\bottomrule
\end{tabular}
\label{table5}
\end{table}

In this subsection, we conduct a comparison of different node information fusion methods based on the embedding of the two channels.

To calculate the attention coefficients, the adaptive attention method relies on the evidence entities $v$ embedding of both graphs, while the shared matrix method does not rely on the node embedding. On the other hand, the fusional attention method relies on the $v$ embedding of both graphs but to different degrees.

As shown in Table \ref{table5}, we observe that the fusional attention method outperforms the shared matrix method and the shared matrix method outperforms adaptive attention, which suggests that the adaptive attention method does not make sensible use of node embedding to calculate the attention coefficients.

In the clinical evidence recommendation task, we aim for higher scores $score(p_i,v_i)$ between entities $p_i$ and $v_i$ for which links exist. In the ECT embedding, $v$ aggregates information about $p$, while in the ETG embedding, $v$ only aggregates information about $v$. When fusing for evidence entities, it is important to consider the downstream tasks fully. Therefore, the best results are achieved with the fusional attention method using $p$ from the ECT embedding as the $query$ vector to calculate the attention coefficients.


\begin{table*}
\centering
\caption{Performance of different similarity}
\begin{tabular}{l|c|c|c|c|c|c|c|c} 
\toprule
$\eta_{thr}$ & NDCG@3         & NDCG@5         & NDCG@7         & NDCG@9         & HR@3           & HR@5           & HR@7           & HR@9            \\ 
\hline\hline
0.0          & 0.281          & 0.288          & 0.296          & 0.300          & \textbf{0.267} & 0.333          & 0.356          & 0.367           \\
0.1          & 0.274          & 0.284          & 0.290          & 0.300          & 0.256          & 0.322          & 0.356          & 0.367           \\
0.2          & 0.263          & 0.285          & 0.288          & 0.300          & 0.256          & 0.322          & 0.356          & 0.367           \\
0.3          & 0.268          & 0.287          & 0.287          & 0.304          & 0.256          & 0.322          & 0.356          & 0.378           \\
0.4          & 0.263          & 0.281          & 0.289          & 0.296          & 0.256          & 0.322          & 0.356          & 0.367           \\
0.5          & 0.272          & 0.278          & 0.287          & 0.292          & 0.256          & 0.322          & 0.344          & 0.378           \\
0.6          & 0.264          & 0.277          & 0.287          & 0.305          & \textbf{0.267} & 0.322          & 0.344          & 0.389           \\
0.7          & 0.259          & 0.280          & 0.290          & 0.297          & 0.256          & 0.333          & 0.367          & 0.389           \\
0.8          & \textbf{0.284} & \textbf{0.319} & \textbf{0.326} & \textbf{0.325} & 0.256          & \textbf{0.344} & \textbf{0.378} & \textbf{0.400}  \\
0.9          & 0.260          & 0.274          & 0.295          & 0.301          & 0.256          & 0.311          & 0.344          & 0.378           \\
1.0          & 0.259          & 0.295          & 0.309          & 0.316          & 0.256          & 0.311          & 0.344          & 0.378           \\
\bottomrule
\end{tabular}
\label{table6}
\end{table*}

\subsection{Robust Study on Similarity}

Table \ref{table6} shows that we perform the embedding of the ETG with a threshold $\eta_{thr}$ ranging from 0 to 1 in steps of 0.1. When $\eta_{thr}=1$, all nodes of ETG only have self-loops, whereas when $\eta_{thr}=0$, all evidence entities are fully connected. Our results indicate that the overall performance is optimal when $\eta_{thr}=0.8$, but our model also performs well at other thresholds. This suggests that our fusional attention is capable of learning the most relevant information when fusing nodes and minimizing the impact of irrelevant information, making MHAN highly robust.


\subsection{Study on Fusional Attention Head}

\begin{figure}[!t]
\centerline{\includegraphics[width=0.9\columnwidth]{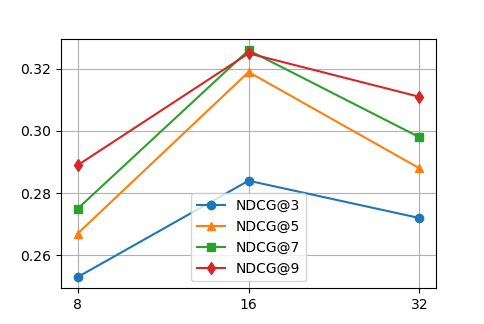}}
\caption{Performance of different fusional attention parameters with NDCG indicators.}
\label{001}
\end{figure}

\begin{figure}[!t]
\centerline{\includegraphics[width=0.9\columnwidth]{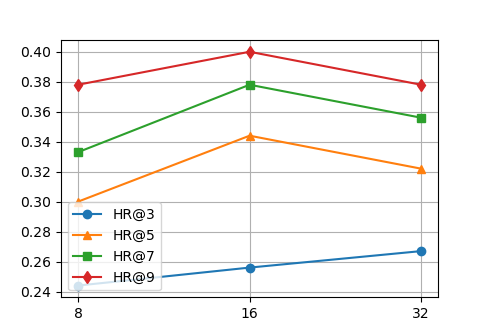}}
\caption{Performance of different fusional attention parameters with HR indicators.}
\label{002}
\end{figure}

We provide a detailed discussion on how the number of fusional attention heads $N$ impacts the model's performance. We demonstrate the influence of parameter $N$ in Fig \ref{001} and \ref{002} by varying it from 8 to 32. From our results, we observe that MHAN attains the optimal performance when $N=16$. This is because an excessive number of attention heads may lead to overfitting, while too few heads may result in insufficient attention and an inability to capture relevant information effectively. 

\subsection{Case Study}

\begin{table}
\centering
\caption{Top five results from COVID-19}
\begin{tabular}{p{0.9\columnwidth} } 
\toprule
\textbf{Problem: }COVID-19  \\ 
\hline
\textbf{NO.1: }A Randomized Controlled Adaptive Study Comparing COVID-19 Convalescent Plasma (CCP) to Non-immune Plasma to Limit Coronavirus-associated Complications in Hospitalized Patients    \\
\textbf{NO.2: }Canakinumab to Reduce Deterioration of Cardiac and Respiratory Function in SARSCoV2 Associated Acute Myocardial Injury With Heightened Inflammation                                \\
\textbf{\textit{NO.3: }}Mavrilimumab to Reduce Progression of Acute Respiratory Failure in Patients With Severe COVID-19 Pneumonia and Systemic Hyper-inflammation                                         \\
\textbf{NO.4: }A Randomized, Double-blind, Placebo-controlled Trial to Evaluate the Efficacy of Hydroxychloroquine and Azithromycin to Prevent Hospitalization or Death in Persons With COVID-19  \\
\textbf{\textit{NO.5: }}Inhaled Dornase Alfa for Treatment of ARDS in Patients With SARS-CoV-2                                                                                                             \\
\bottomrule
\end{tabular}
\label{table7}
\end{table}

\begin{table}
\centering
\caption{Top four results from \\ Refractory Malignant Solid Neoplasm}
\begin{tabular}{p{0.9\columnwidth} } 
\toprule
\textbf{Problem: } Refractory Malignant Solid Neoplasm                                                                                                                                                                           \\ 
\hline
\textbf{NO.1: }A MATCH Treatment Subprotocol Y: AZD5363 in Patients With Tumors With AKT Mutations    \\
\textbf{\textit{NO.2: }}BN-Brachyury, Entinostat, Adotrastuzumab Emtansine and M7824 in Advanced Stage Breast Cancer                                \\
\textbf{NO.3: }MATCH Treatment Subprotocol Z1A: Binimetinib in Patients With Tumors (Other Than Melanoma) With NRAS Mutations                                         \\
\textbf{NO.4: }MATCH Treatment Subprotocol W: Phase II Study of AZD4547 in Patients With Tumors With Aberrations in the FGFR Pathway  \\

\bottomrule
\end{tabular}
\label{table8}
\end{table}

We present various cases of recommendation results to analyze the effectiveness of our model. Table \ref{table7} displays the top five evidence titles recommended by MHAN for the problem COVID-19. Among these, evidence No.1, 2, and 4, which have the problem COVID-19, belong to the test set, while evidence No.3 and 5 are not part of it and do not involve COVID-19 in their problems. Our model can accurately recommend the three pieces of evidence in the test set because their titles and descriptive information are similar to those in the training set, and their interconnection in the ETG facilitates message passing. Furthermore, the model recommends two evidence pieces that are not in the test set but relevant to COVID-19 from their titles. The titles of evidence No.3 and 5 indicate their medical relevance to the study of COVID-19 since SARS-CoV-2 is the virus that causes it. Despite their problems not being COVID-19, our model is still able to identify them. We further investigated the problems of these two evidence pieces, and found that NO.3 is related to pneumonia and SARS-CoV 2, while NO.5 is related to ARDS and SARS-CoV 2.

Table \ref{table8} presents another case where we analyze the effectiveness of our model's recommendations. Evidence No.1, 3, and 4 with the problem Refractory Malignant Solid Neoplasm are in the test set, while evidence No.2 is not. Despite this, our model recommends evidence No.2 because its research outcome, descriptive information, or intervention may have similarities with the problem. This recommendation highlights the potential for our model to inspire the discovery of new evidence for Refractory Malignant Solid Neoplasm.


\section{Conclusion}
In this paper, we explore the clinical evidence recommendation task, which predicts a set of evidence given a problem. To alleviate the sparsity, we construct ETG based on linguistic similarity. We solve the co-reference-text heterogeneity using dual channels and demonstrate experimentally that linguistic linking is not equivalent to traditional edge heterogeneity. In order to leverage the textual information of the evidence entities, we also use the textual information of the evidence entities as the initial embedding vectors of the graphs. In order to fuse the information of nodes aggregated from different channels, we propose fusional attention. We prove the advantages of MHAN over state-of-art methods.

There are several directions for feature work based on our research. A meaningful direction would be to extract richer entities based on textual information from publicly available datasets to enrich the graph representation. Our research also has practical implications, and although it is still a little far from practical application in clinical medical treatment, it can guide the direction of potential evidence or drug research in evidence-based medicine.

\end{document}